\documentclass{article}

\usepackage[preprint]{neurips_2024}

\usepackage[utf8]{inputenc} 
\usepackage[T1]{fontenc}    
\usepackage{hyperref}       
\usepackage{url}            
\usepackage{booktabs}       
\usepackage{amsfonts}       
\usepackage{nicefrac}       
\usepackage{microtype}      
\usepackage[table]{xcolor}         
\definecolor{Gray}{gray}{0.7}
\usepackage{listings}
\usepackage{graphicx}
\usepackage{subcaption}

\title{Dubo-SQL: Diverse Retrieval-Augmented Generation and Fine Tuning for Text-to-SQL}

\author{%
  Dayton G. Thorpe \\
  Mercator Technologies \\
  Palo Alto, CA \\
  \texttt{dayton@mercator.tech} \\
  \And
  Andrew J. Duberstein \\
  Mercator Technologies \\
  San Francisco, CA \\
  \texttt{ajduberstein@gmail.com} \\
  \And
  Ian A. Kinsey \\
  Mercator Technologies \\
  New York, NY \\
  \texttt{ian@mercator.tech}
}

\begin{document}

\maketitle

\begin{abstract}
The current state-of-the-art (SOTA) for automated text-to-SQL still falls well short of expert human performance as measured by execution accuracy (EX) on the BIRD-SQL benchmark. The most accurate methods are also slow and expensive. To advance the SOTA for text-to-SQL while reducing cost and improving speed, we explore the combination of low-cost fine tuning, novel methods for diverse retrieval-augmented generation (RAG) and new input and output formats that help large language models (LLMs) achieve higher EX. We introduce two new methods, Dubo-SQL v1 and v2. Dubo-SQL v1 sets a new record for EX on the holdout test set of BIRD-SQL. Dubo-SQL v2 achieves even higher performance on the BIRD-SQL dev set. Dubo-SQL v1 relies on LLMs from OpenAI, but uses the low-cost GPT-3.5 Turbo while exceeding the performance of the next-best model using OpenAI, which instead uses the more expensive GPT-4. Dubo-SQL v1 exceeds the performance of the next-best model using GPT-3.5 by over 20\%. Dubo-SQL v2 uses GPT-4 Turbo and RAG in place of fine tuning to push EX higher.
\end{abstract}

\section{Introduction}

Text-to-SQL models and pipelines are rapidly improving, but still lag far behind human performance. Measured by execution accuracy on the BIRD-SQL \citep{BIRD} holdout test set, advancements in Large Language Models (LLMs) have pushed forward the state-of-the-art from 12.89\% with T5-Base to 54.89\% with GPT-4 with simple zero-shot prompts. The gap between GPT-4’s zero-shot performance and human performance of 92.96\% leaves an important open area for research. 

Compared to code-completion and code-writing tasks in languages like Python or C++, writing SQL is more dependent on extensive knowledge of a given user's context. The full context required to write accurate SQL can itself be millions of words. In Python, many functions can be written without any user-specific context at all, and most others can be written with only concise documentation for the input and output of a small number of related functions. Answering questions about business metrics in SQL, such as "what is my contribution margin?", requires knowing both how the users define their metrics and which tables and columns contain the necessary information. Large corporations commonly have databases with thousands to hundreds of thousands of tables along with natural language documentation that can be longer than the database schemas themselves.  Without access to the metrics definitions, the database schema, and the related documentation, a text-to-SQL model can, at best, write syntactically valid SQL while hallucinating the definition of the user's metric and the tables and columns needed to compute it.

Research into how LLMs can be coaxed into writing more accurate code to answer more complex questions informs how we advance text-to-SQL execution accuracy. Few-shot learning \citep{Brown2020}, which gives a small number of question-answer pairs as part of the system or user prompt without any change to model weights, can approach or exceed the performance of fine-tuned models at much lower upfront cost. Chain-of-Thought prompting \citep{wei2023chainofthought} enables LLMs to answer more complex questions accurately by using additional output tokens to discuss intermediate steps and mimic human reasoning. Self-reflection \citep{Shinn2023} calls an external service to check the LLM's answer, and then has the LLM discuss the feedback from the external service and update its answer. Many researchers have helped to develop autonomous agents \citep{wang2024survey} which can call external functions or additional LLMs as sub-agents and self-determine when they have successfully answered a question or completed a task. In a prescribed domain like text-to-SQL, a predetermined plan can place multiple calls to LLMs and other functions to break the problem into easier components, without granting full autonomy to the LLM to decide when the task is complete.

Predating these techniques, fine tuning \citep{devlin-etal-2019-bert} allows a general-purpose LLM to be specialized for a specific task given as few as hundreds of labeled training examples. Fine tuning requires many more examples than few-shot learning, which can bring material improvements in performance with even a single example, but many fewer examples than LLM pre-training, which relies on large web corpora. Although fine tuning comes with an upfront cost, it can save money at inference time in two ways. First, a fine-tuned small model may exceed the performance of a larger model that has not been fine tuned. Second, fine tuning can replace both a system prompt and few shot examples, so new questions can be submitted with a more concise prompt. If the the number of times the model will be used for inference is high enough, the total cost of fine tuning and inference together can be lower than using few-shot prompting.

In this paper, we introduce two new text-to-SQL methods. The first, Dubo-SQL v1, is a low-cost, token-efficient fine tuning method that set a new record on the BIRD-SQL benchmark at the time of its submission and at lower cost than its nearest competitors. The second, Dubo-SQL v2, uses a novel retrieval-augmented generation pipeline to exceed the performance of Dubo-SQL v1 when used with GPT-4 Turbo. Our code is available at our GitHub repository \footnote{\url{https://github.com/mercatorhq/dubo-sql}}.

\section{Related Work}

\citet{gao2023dailsql} systematically explore LLM-based methods for text-to-SQL and combine the best techniques to create DAIL-SQL. In particular, they consider 1) five options for formatting the database schema and user question, 2) four options for selecting the most relevant few-shot examples, and 3) two options for formatting the few-shot examples. They synthesize best practices and introduce sophisticated new methods – DAIL Selection and DAIL Organization – for selecting and formatting few-shot examples. We take inspiration from this work while choosing simpler methods.

\citet{pourreza2024dinsql} create DIN-SQL to decompose the full text-to-SQL problem into simpler sub-tasks chained together for better overall execution accuracy. We build upon that work here, though we use at most two modules: one for a first draft of the SQL, and another for correcting syntax errors reported by the compiler. 

One key component of both versions of Dubo-SQL is using a SQL compiler to execute the SQL generated by the LLM. If the compiler generates an error, the SQL and the error message are sent back to the LLM for correction. MAC-SQL \citep{wang2024macsql} similarly uses what the authors term a "refiner" to correct SQL syntax errors. Dubo-SQL and DIN-SQL both use similar methods and were submitted to the BIRD leader board before MAC-SQL was submitted or published.

\section{Methodology}

We introduce Dubo-SQL v1, a low-cost fine-tuning approach that achieved record performance on BIRD-SQL when combined with GPT-3.5 Turbo, and Dubo-SQL v2, a RAG-based method built on GPT-4 Turbo that achieves even higher performance with a longer context window and without fine tuning. The longer context window of GPT-4 Turbo is unnecessary for BIRD, but is crucial for answering questions about corporate databases that often contain thousands of tables.

For all OpenAI API calls, we use a temperature of 0, except when explicitly stated otherwise, and specify a random seed, but we still see residual fluctuations in behavior given the same input only moments apart. Specifying the seed reduces the variation somewhat. Running the same input on the same sample of 200 questions from the BIRD dev set, without specifying a random seed we found that the execution accuracy varied by 1.5\% over three runs. When we specified the same random seed on three runs under the same conditions, the execution accuracy varied by 0.5\%. All reported comparisons use the same random seed.

\subsection{Dubo-SQL v1}

First, we prepare the table schema and sample data in the format shown in Appendix Listing-\ref{lst:schema}, with a specific example given in Appendix Listing-\ref{lst:example}. We include up to five rows of sample data, reducing the number of rows when the number of tokens per row causes the entire prompt to exceed the limits of the model’s context window. Foreign keys are included, if provided in the database, but primary keys and column constraints are deleted, with the goal of providing only the information a human data analyst would typically use when writing a SQL query.

We then assemble the prompt in the format shown in Appendix Listing-\ref{lst:usrprompt}, with a specific example given in Appendix Listing-\ref{lst:usrpromptex}. We choose a prompt that is as concise and direct as possible. We do not include any examples or demonstrations, a system prompt, a role assignment, e.g. “You are a senior data analyst”, a task description, or instructions. We provide the database context without explanation and end by asking the user’s question.

Finally, we submit the BIRD training set examples to OpenAI for fine tuning on GPT-3.5 Turbo. The ideal responses used for fine tuning directly state the SQL query without any introduction, chain of thought, or surrounding formatting such as JSON or a code block. We train for 2 epochs with the 9,424 BIRD training examples, using 17 million tokens.

For testing, prompts are constructed in the same manner and submitted to the fine-tuned model with a maximum token count of 1,000. In all cases, we include the evidence, explained further below, provided by BIRD. Before finalizing the answer, we attempt to execute the SQL. If it returns an error message, we submit a second request to the same fine-tuned model with the original user prompt and response, followed by a second user prompt sharing the error and asking for corrected SQL, as shown in Appendix Listing-\ref{lst:correctionprompt}.

\begin{figure}[htbp] 
    \centering
    \includegraphics[width=\linewidth]{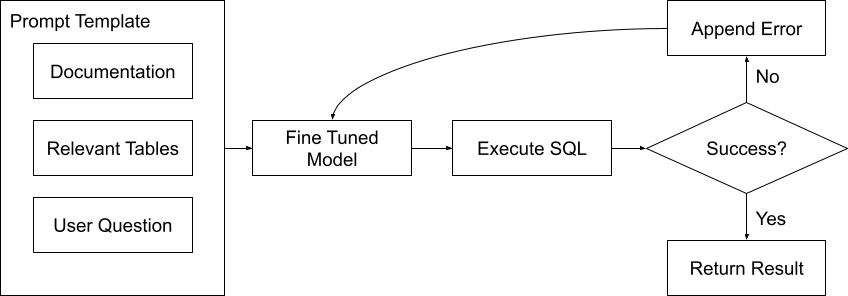}
    \caption{A diagram of Dubo-SQL v1} 
    \label{fig:dubov1} 
\end{figure}

\subsection{Dubo-SQL v2}

For Dubo-SQL v2, instead of fine tuning, we use a system prompt and few-shot examples. The system prompt, given in Appendix Listing-\ref{lst:systemprompt}, includes a role assignment and descriptions of the task and input and output formats. 

In Dubo-SQL v1, we presented sample data in a CSV format. For Dubo-SQL v2, we use a more LLM-readable input format. Whereas traditional data analysis tools like Python Pandas can easily read CSV data and humans can easily read CSV data by moving their eyes up and down a given column, LLMs may struggle to understand which data point goes with which column in a long CSV. Instead, we list each column on its own line, along with any available column description and up to five example values. The system prompt in Appendix Listing-\ref{lst:systemprompt} includes a template of the database schema format. This schema format is similar to that in MAC-SQL, with minor modifications, while the rest of the prompt and pipeline is significantly different.

Other improvements we introduce are:

\begin{itemize}
\item \textbf{Diverse RAG} To select few-shot examples, we start by calculating the vector embeddings of all natural language questions in the BIRD training data set using OpenAI’s text-embedding-3-large. We embed only the question, not including the evidence or ground truth SQL. For each new question, we select few-shot examples based on cosine similarity. We find that in many cases, the most similar questions are only minor variations on each other. In order to obtain more informative examples, we only allow one example question-answer pair for any single reference database.

\item \textbf{Few-shot examples in conversation history} Prior work included few-shot examples either in the system prompt or in the user prompt with an explicit statement that they are meant to be examples for the LLM to learn from. We find the LLM learns more from the few-shot examples if we include the questions as user messages and the answers as assistant messages in the conversation history, all in the same format for the new user question and the expected format for the LLM's response.

\item \textbf{JSON output} We find that asking for the output in JSON format improves execution accuracy versus asking for the SQL to be given directly, without a surrounding code block or JSON.
\end{itemize}

\begin{figure}[htbp] 
    \centering
    \includegraphics[width=\linewidth]{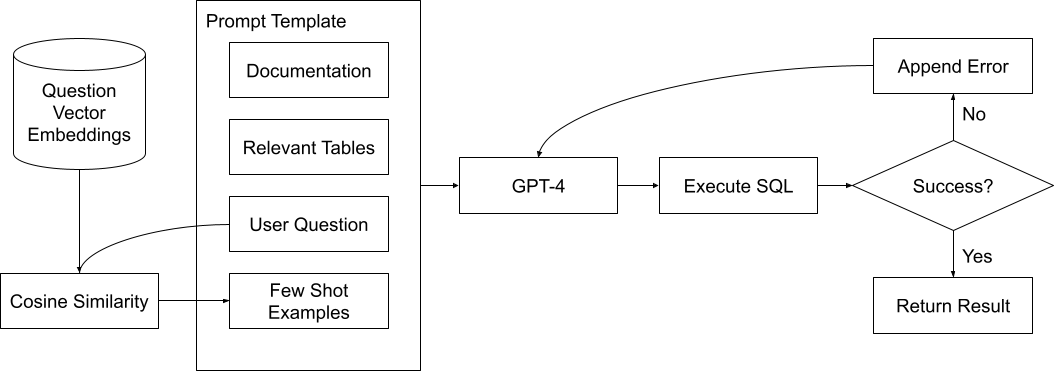}
    \caption{A diagram of Dubo-SQL v2} 
    \label{fig:dubov2} 
\end{figure}

\section{Experiment}

\subsection{Models}

For Dubo-SQL v1, we create our own fine-tune of gpt-3.5-turbo-0613. Dubo-SQL v2 uses gpt-4-0125-preview without fine tuning.

\subsection{Dataset}

We test our approach on the BIRD-SQL dataset. For our fine-tuned model, our dev set results come from a model fine-tuned on the training set only and our results on the test set, which is held back from the public, come from a model further fine-tuned on the dev set. The BIRD dataset establishes a new standard in text-to-SQL benchmarking, with real-world user questions asked of 95 databases pulled from 37 industries. Each user question is paired with additional information, labeled “evidence”, explaining how terms used in the user question relate to the database. For example, “NBA refers to lgID = 'NBA'”.

\subsection{Metrics}

We evaluate performance using execution accuracy (EX) as defined by BIRD. EX compares the results of executing the ground truth SQL and the predicted SQL against the associated database as sets. If the predicted SQL is different from the ground truth SQL but returns the same data, it is marked as correct.

We also consider the cost of training and inference. In production settings, the cost of building and running a model is crucial to its usefulness.

\section{Results}

\subsection{Execution Accuracy}

On the BIRD test set, Dubo-SQL v1 achieves a test set execution accuracy of 60.71\%. This compares to similarly constructed models, built on top of OpenAI’s proprietary models, MAC-SQL with 59.59\%, DAIL-SQL with 57.41\%, and DIN-SQL with 55.90\%, as well as the zero-shot GPT-4 baseline provided by the BIRD team of 54.89\%. Our model, built on top of GPT-3.5 Turbo, beats the ChatGPT baseline of 39.30\% by 21.41\%.  When access to GPT-4 fine tuning becomes publicly available, repeating the procedure for training Dubo-SQL v1 may bring significant further improvements in execution accuracy. See the full comparison of Dubo-SQL v1 to other models in Table \ref{tab:test_set_accuracy}.

\begin{table}[htbp]
  \centering
  \caption{Execution accuracy on the holdout test set of BIRD-SQL for the top 11 models retrieved from \url{https://bird-bench.github.io/} on December 3, 2023.}
  \begin{tabular}{lc}
    \toprule
    \rowcolor{Gray}
    \textbf{Method} & \textbf{EX (\%)} \\
    \midrule
    \addlinespace[0.25em]
    Dubo-SQL v1 (ours) & 60.71 \\
    SFT CODES-15B \citep{li2024codes} & 60.37 \\
    MAC-SQL + GPT-4 \citep{wang2024macsql} & 59.59 \\
    SFT Code-S-7B \citep{li2024codes} & 59.25 \\
    DAIL-SQL + GPT-4 \citep{gao2023dailsql} & 57.41 \\
    DIN-SQL + GPT-4 \citep{pourreza2024dinsql} & 55.90 \\
    GPT-4 \citep{BIRD} & 54.89 \\
    Claude-2 \citep{BIRD} & 49.02 \\
    Open-SQL & 47.74 \\
    ChatGPT + CoT \citep{BIRD} & 40.08 \\
    ChatGPT \citep{BIRD} & 39.30 \\
    \bottomrule
  \end{tabular}
  \label{tab:test_set_accuracy}
\end{table}

On the BIRD-SQL dev set, Dubo-SQL v2 has an execution accuracy of 61.47\%, 1.63\% ahead of the Dubo-SQL v1 performance on the same set, though still behind MCS-SQL ad GRA-SQL. The diverse few-shot examples effectively replace the fine tuning step with even higher execution accuracy when combined with an LLM with better baseline reasoning abilities. At the trade-off of higher inference costs, Dubo-SQL v2 has a lower setup cost than Dubo-SQL v1. Given a new context, such as a large corporate database, it could offer a faster path to early testing. See the full comparison of Dubo-SQL v2 to other models in Table \ref{tab:dev_set_accuracy}.

\begin{table}[htbp]
  \centering
  \caption{Execution accuracy on the dev set of BIRD-SQL for Dubo-SQL v2 and publicly reported results from other models, retrieved from \url{https://bird-bench.github.io/} on April 10, 2024.}
  \label{tab:dev_set_accuracy}
  \begin{tabular}{lc}
    \toprule
    \rowcolor{Gray}
    \textbf{Method} & \textbf{EX (\%)} \\
    \midrule
    \addlinespace[0.25em]
    MCS-SQL + GPT-4 & 63.36 \\
    GRA-SQL & 62.58 \\
    Dubo-SQL v2 + GPT-4 (ours) & 61.47 \\
    OpenSearch-SQL\_v1 + GPT-4 & 61.34 \\
    PB-SQL, v1 & 60.50 \\
    Dubo-SQL v1 (ours) & 59.71 \\
    SFT CodeS-15B \citep{li2024codes} & 58.47 \\
    \{Chat2Query\} (GPT-4 + data entity modeling) & 58.15 \\
    (PingCAP) & \\
    DTS-SQL + DeepSeek 7B \citep{pourreza2024dtssql} & 55.8 \\
    SENSE & 55.48 \\
    \bottomrule
  \end{tabular}
\end{table}

\subsection{Cost}

For Dubo-SQL v1, training cost \$273 at OpenAI’s pricing of \$8.00 per million training tokens. With our concise prompt format, the median and 95th percentile total token counts per question are 1686 and 3,327, including a median and 95th percentile of 49 and 80 output tokens. At OpenAI's pricing for fine-tuned models of \$3 per million input tokens and \$6 per million output tokens, that gives an inference cost under \$0.01 per natural language question at the 95th percentile. This cost compares to the reported \$0.50 for each question using DIN-SQL, which achieves an EX 4.81\% absolute lower than Dubo-SQL v1. With 1,533 questions in the dev set alone, the combined inference and training costs are lower for Dubo-SQL v1 than for DIN-SQL. In a production environment with an even higher ratio of inference to training questions, the cost savings will be greater.

For Dubo-SQL v2, the median and 95th percentile token counts per natural language question are 7,970 and 13,599. At OpenAI's pricing of \$10.00 per million input tokens and \$30 per million output tokens for GPT-4 Turbo, the inference cost is under \$0.14 per question, far above the cost of Dubo-SQL v1 but still well below the cost of DIN-SQL.

\subsection{Ablation Study}
We evaluated Dubo-SQL v2 with and without each of the improvements we introduced, using a consistent random sample of 500 questions from the BIRD dev set. The error correction loop adds 2.2\% EX. Requesting the output in JSON format, instead of plain text, also adds 2.2\% EX. Switching from simple retrieval-augmented generation that uses the most similar questions to our method for selecting a diverse set of similar questions adds 2.6\% EX.

Our main results come from combining our prompting methods with GPT-4 Turbo. Using the same fine-tuned model from Dubo-SQL v1 causes a regression of 6.6\% and using a non-fine-tuned GPT-3.5 Turbo causes a 16.0\% regression in EX.  See the results in Table \ref{tab:ablation_study}

\begin{figure}[htbp]
    \centering
    \begin{subfigure}{0.48\textwidth}
        \centering
        \includegraphics[width=\linewidth]{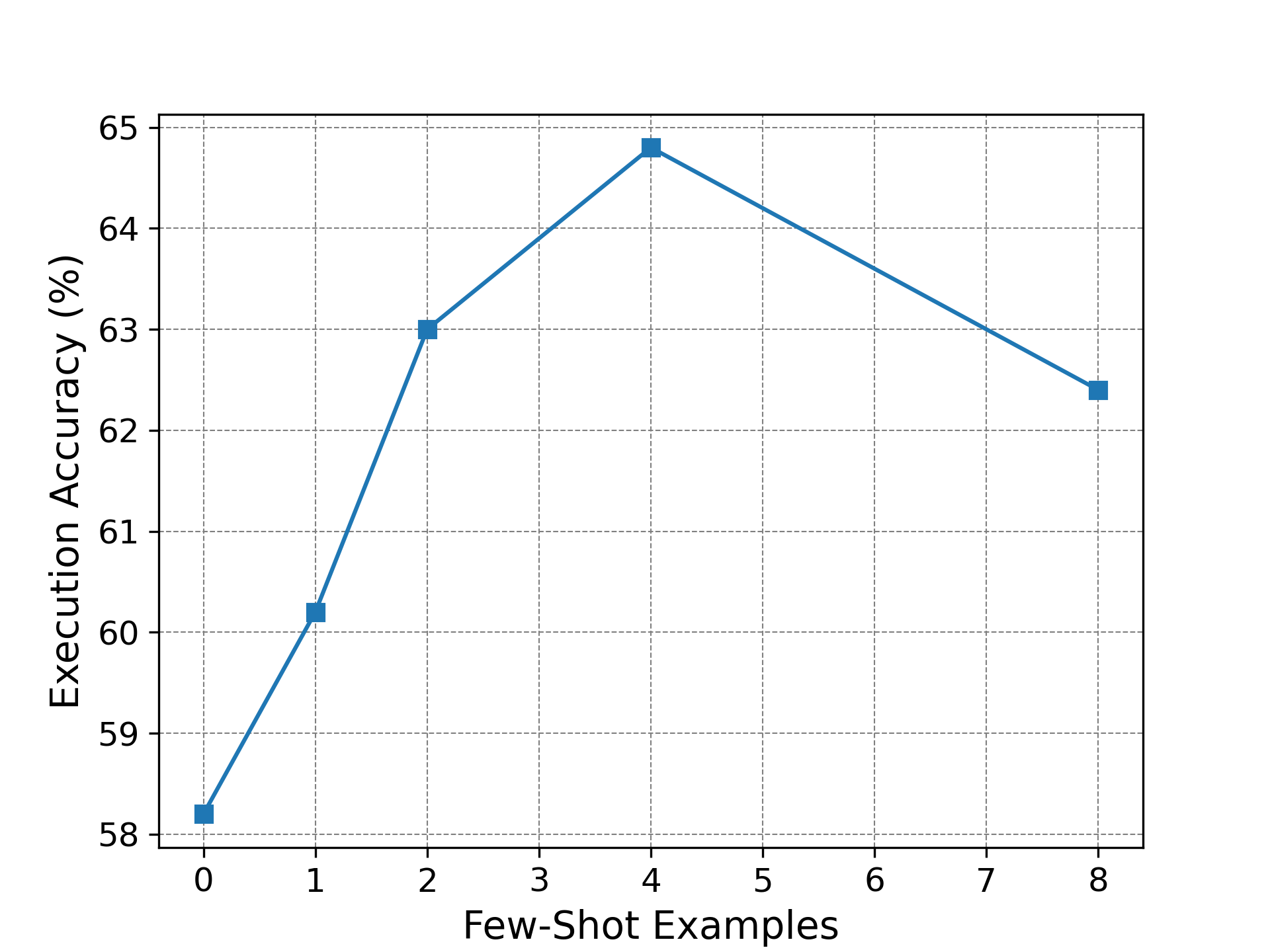}
    \end{subfigure}
    \begin{subfigure}{0.48\textwidth}
        \centering
        \includegraphics[width=\linewidth]{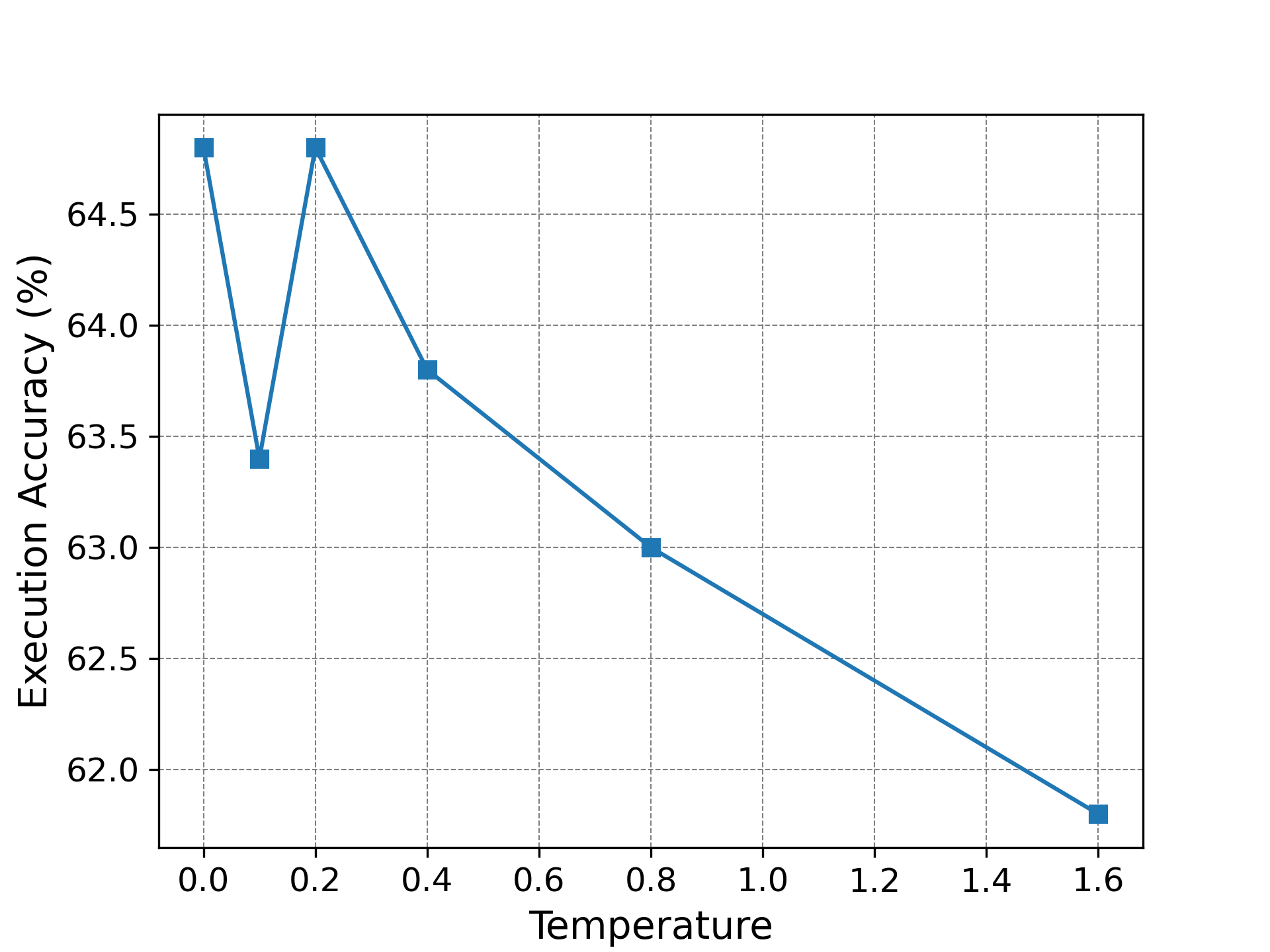}
    \end{subfigure}
    \caption{a) Performance with varying numbers of few-shot examples. b) Performance with non-zero temperature}
    \label{fig:parameters}
\end{figure}

We evaluated running the model at a non-zero temperature, but found that it only hurt performance, so we continued using a temperature of 0 in Dubo-SQL v2 (see Fig-\ref{fig:parameters}). We also explored using one to eight few-shot examples and found four to be ideal (see Fig-\ref{fig:parameters}). We expect the ideal number of few-shot examples would vary by LLM.

\begin{table}[htbp]
  \centering
  \caption{Execution Accuracy of Dubo-SQL v2 ablation study using a consistent sample of 500 question-answer pairs from the BIRD-SQL dev set.}
  \label{tab:ablation_study}
  \begin{tabular}{lc}
    \toprule
    \rowcolor{Gray}
    \textbf{Method} & \textbf{EX (diff from Dubo-SQL v2) (\%)} \\
    \midrule
    \addlinespace[0.25em]
    Dubo-SQL v2 + GPT-4 Turbo & 64.8 \\
    \hspace*{1em}w/o error correction & 62.6 (-2.2) \\
    \hspace*{1em}w/o JSON output & 62.6 (-2.2) \\
    \hspace*{1em}w/ non-diverse RAG & 62.2 (-2.6) \\
    \hspace*{1em}w/ fine-tuned model from Dubo-SQL v1 & 58.2 (-6.6) \\
    \hspace*{1em}w/ GPT-3.5 Turbo & 48.8 (-16.0) \\
    \bottomrule
  \end{tabular}
\end{table}

\section{Conclusions}
Advancements in the size and architecture of LLMs have dramatically improved their performance on text-to-SQL. Previous work to decompose the text-to-SQL problem into simpler sub-problems and to select the most relevant few-shot examples have lifted performance further above the baseline zero-shot case. In this paper, we have introduced a low-cost fine-tuning approach to exceed the performance of these methods with dramatically lower inference costs. We have also introduced new input and output formats that raise LLM performance higher and improved the selection process for few-shot examples. By showing the model a diverse set of example questions and presenting them in the conversation history instead of the user- or system-prompt, we help the LLM learn more in context.

\section{Limitations}
Although the BIRD-SQL dataset uses real world databases and user questions, the databases are still small relative to common corporate databases. BIRD databases can contain dozens of tables with tens of columns each, while corporate databases commonly contain thousands of tables with hundreds of columns each. Including all table schemas and five rows of sample data for such databases could require millions of tokens. With models available today, an additional step is required to select a short list of relevant tables from a much longer list that exceeds the context limits of most open-source or proprietary LLMs. Dubo-SQL v1 uses a fine-tuned version of gpt-3.5-turbo-0613, which has a context window of 4,096 tokens. Dubo-SQL v2, built with gpt-4-0125-preview, has a context limit of 128,000 tokens. That limit is more than sufficient to contain table schemas and five rows of sample data for all databases in the BIRD train and dev sets, but is insufficient for large corporate databases. This context window is too short to combine our fine-tuned model with few-shot examples. In future work, we may combine few-shot examples with fine-tuning on a model such as gpt-3.5-turbo-0125, which has a longer context window of 16,385 tokens. Inference with Dubo-SQL v2 is significantly more expensive than with Dubo-SQL v1 because it uses more tokens and a larger model. Fine tuning gpt-3.5-turbo-0125 may be able to match the execution accuracy of Dubo-SQL v2 at lower cost by using a smaller model and eliminating the system prompt.

\begin{ack}
This work was funded by Mercator Technologies. The authors are co-founders of Mercator Technologies. We wish to thank the authors of BIRD for running our code on their holdout test set.
\end{ack}

\bibliographystyle{plainnat}
\bibliography{dubo_sql}

\begin{thebibliography}{11}
\providecommand{\natexlab}[1]{#1}
\providecommand{\url}[1]{\texttt{#1}}
\expandafter\ifx\csname urlstyle\endcsname\relax
  \providecommand{\doi}[1]{doi: #1}\else
  \providecommand{\doi}{doi: \begingroup \urlstyle{rm}\Url}\fi

\bibitem[Brown et~al.(2020)Brown, Mann, Ryder, Subbiah, Kaplan, Dhariwal, Neelakantan, Shyam, Sastry, Askell, Agarwal, Herbert-Voss, Krueger, Henighan, Child, Ramesh, Ziegler, Wu, Winter, Hesse, Chen, Sigler, Litwin, Gray, Chess, Clark, Berner, McCandlish, Radford, Sutskever, and Amodei]{Brown2020}
Tom Brown, Benjamin Mann, Nick Ryder, Melanie Subbiah, Jared~D Kaplan, Prafulla Dhariwal, Arvind Neelakantan, Pranav Shyam, Girish Sastry, Amanda Askell, Sandhini Agarwal, Ariel Herbert-Voss, Gretchen Krueger, Tom Henighan, Rewon Child, Aditya Ramesh, Daniel Ziegler, Jeffrey Wu, Clemens Winter, Chris Hesse, Mark Chen, Eric Sigler, Mateusz Litwin, Scott Gray, Benjamin Chess, Jack Clark, Christopher Berner, Sam McCandlish, Alec Radford, Ilya Sutskever, and Dario Amodei.
\newblock Language models are few-shot learners.
\newblock In H.~Larochelle, M.~Ranzato, R.~Hadsell, M.F. Balcan, and H.~Lin, editors, \emph{Advances in Neural Information Processing Systems}, volume~33, pages 1877--1901. Curran Associates, Inc., 2020.
\newblock URL \url{https://proceedings.neurips.cc/paper_files/paper/2020/file/1457c0d6bfcb4967418bfb8ac142f64a-Paper.pdf}.

\bibitem[Devlin et~al.(2019)Devlin, Chang, Lee, and Toutanova]{devlin-etal-2019-bert}
Jacob Devlin, Ming-Wei Chang, Kenton Lee, and Kristina Toutanova.
\newblock {BERT}: Pre-training of deep bidirectional transformers for language understanding.
\newblock In Jill Burstein, Christy Doran, and Thamar Solorio, editors, \emph{Proceedings of the 2019 Conference of the North {A}merican Chapter of the Association for Computational Linguistics: Human Language Technologies, Volume 1 (Long and Short Papers)}, pages 4171--4186, Minneapolis, Minnesota, June 2019. Association for Computational Linguistics.
\newblock \doi{10.18653/v1/N19-1423}.
\newblock URL \url{https://aclanthology.org/N19-1423}.

\bibitem[Gao et~al.(2023)Gao, Wang, Li, Sun, Qian, Ding, and Zhou]{gao2023dailsql}
Dawei Gao, Haibin Wang, Yaliang Li, Xiuyu Sun, Yichen Qian, Bolin Ding, and Jingren Zhou.
\newblock Text-to-sql empowered by large language models: A benchmark evaluation, 2023.

\bibitem[Li et~al.(2024{\natexlab{a}})Li, Zhang, Liu, Fan, Zhang, Zhu, Wei, Pan, Li, and Chen]{li2024codes}
Haoyang Li, Jing Zhang, Hanbing Liu, Ju~Fan, Xiaokang Zhang, Jun Zhu, Renjie Wei, Hongyan Pan, Cuiping Li, and Hong Chen.
\newblock Codes: Towards building open-source language models for text-to-sql, 2024{\natexlab{a}}.

\bibitem[Li et~al.(2024{\natexlab{b}})Li, Hui, Qu, Yang, Li, Li, Wang, Qin, Geng, Huo, et~al.]{BIRD}
Jinyang Li, Binyuan Hui, Ge~Qu, Jiaxi Yang, Binhua Li, Bowen Li, Bailin Wang, Bowen Qin, Ruiying Geng, Nan Huo, et~al.
\newblock Can llm already serve as a database interface? a big bench for large-scale database grounded text-to-sqls.
\newblock \emph{Advances in Neural Information Processing Systems}, 36, 2024{\natexlab{b}}.

\bibitem[Pourreza and Rafiei(2024{\natexlab{a}})]{pourreza2024dinsql}
Mohammadreza Pourreza and Davood Rafiei.
\newblock Din-sql: Decomposed in-context learning of text-to-sql with self-correction.
\newblock \emph{Advances in Neural Information Processing Systems}, 36, 2024{\natexlab{a}}.

\bibitem[Pourreza and Rafiei(2024{\natexlab{b}})]{pourreza2024dtssql}
Mohammadreza Pourreza and Davood Rafiei.
\newblock Dts-sql: Decomposed text-to-sql with small large language models, 2024{\natexlab{b}}.

\bibitem[Shinn et~al.(2023)Shinn, Cassano, Gopinath, Narasimhan, and Yao]{Shinn2023}
Noah Shinn, Federico Cassano, Ashwin Gopinath, Karthik Narasimhan, and Shunyu Yao.
\newblock Reflexion: language agents with verbal reinforcement learning.
\newblock In A.~Oh, T.~Neumann, A.~Globerson, K.~Saenko, M.~Hardt, and S.~Levine, editors, \emph{Advances in Neural Information Processing Systems}, volume~36, pages 8634--8652. Curran Associates, Inc., 2023.
\newblock URL \url{https://proceedings.neurips.cc/paper_files/paper/2023/file/1b44b878bb782e6954cd888628510e90-Paper-Conference.pdf}.

\bibitem[Wang et~al.(2024{\natexlab{a}})Wang, Ren, Yang, Liang, Bai, Chai, Yan, Zhang, Yin, Sun, and Li]{wang2024macsql}
Bing Wang, Changyu Ren, Jian Yang, Xinnian Liang, Jiaqi Bai, Linzheng Chai, Zhao Yan, Qian-Wen Zhang, Di~Yin, Xing Sun, and Zhoujun Li.
\newblock Mac-sql: A multi-agent collaborative framework for text-to-sql, 2024{\natexlab{a}}.

\bibitem[Wang et~al.(2024{\natexlab{b}})Wang, Ma, Feng, Zhang, Yang, Zhang, Chen, Tang, Chen, Lin, et~al.]{wang2024survey}
Lei Wang, Chen Ma, Xueyang Feng, Zeyu Zhang, Hao Yang, Jingsen Zhang, Zhiyuan Chen, Jiakai Tang, Xu~Chen, Yankai Lin, et~al.
\newblock A survey on large language model based autonomous agents.
\newblock \emph{Frontiers of Computer Science}, 18\penalty0 (6):\penalty0 1--26, 2024{\natexlab{b}}.

\bibitem[Wei et~al.(2022)Wei, Wang, Schuurmans, Bosma, Xia, Chi, Le, Zhou, et~al.]{wei2023chainofthought}
Jason Wei, Xuezhi Wang, Dale Schuurmans, Maarten Bosma, Fei Xia, Ed~Chi, Quoc~V Le, Denny Zhou, et~al.
\newblock Chain-of-thought prompting elicits reasoning in large language models.
\newblock \emph{Advances in neural information processing systems}, 35:\penalty0 24824--24837, 2022.

\end{thebibliography}

\newpage
\appendix
\section*{Appendix}
\begin{lstlisting}[label={lst:schema}, caption={Table schema template for Dubo-SQL v1}, breaklines=true]
<table> (<column> <data_type>,..., <column> <data_type>, FOREIGN KEY <column> REFERENCES <table> <column>)
INSERT INTO <table_name> VALUES
(<row_1>)
...
(<row_5>);
\end{lstlisting}

\begin{lstlisting}[label={lst:example}, caption={Specific example of table schema for Dubo-SQL v1}]
CBSA (CBSA INT, CBSA_name TEXT, CBSA_type TEXT);
INSERT INTO CBSA VALUES
(10300, 'Adrian, MI', 'Micro')
(10380, 'Aguadilla-Isabela, PR', 'Metro')
(10420, 'Akron, OH', 'Metro')
(10500, 'Albany, GA', 'Metro')
(10580, 'Albany-Schenectady-Troy, NY', 'Metro');
\end{lstlisting}

\begin{lstlisting}[label={lst:usrprompt}, caption={User prompt format for Dubo-SQL v1}]
<table_schemas>

## The user has asked:
<question>
Note: <evidence>
\end{lstlisting}

\begin{lstlisting}[label={lst:usrpromptex}, caption={User prompt format for Dubo-SQL v1}, breaklines=true]
awards_players (playerID TEXT, award TEXT, year INT, lgID TEXT, note TEXT, pos TEXT, FOREIGN KEY playerID REFERENCES players playerID);
INSERT INTO awards_players VALUES
('abdulka01', 'All-Defensive Second Team', 1969, 'NBA', NULL, NULL)
('abdulka01', 'All-NBA Second Team', 1969, 'NBA', NULL, 'C')
('abdulka01', 'Rookie of the Year', 1969, 'NBA', NULL, NULL)
('abdulka01', 'All-Defensive Second Team', 1970, 'NBA', NULL, NULL)
('abdulka01', 'All-NBA First Team', 1970, 'NBA', NULL, 'C');
...
"series_post" (id INT, year INT, round TEXT, series TEXT, tmIDWinner TEXT, lgIDWinner TEXT, tmIDLoser TEXT, lgIDLoser TEXT, W INT, L INT, FOREIGN KEY tmIDWinner, year REFERENCES teams tmID, year, FOREIGN KEY tmIDLoser, year REFERENCES teams tmID, year);
INSERT INTO "series_post" VALUES
(1, 1946, 'F', 'O', 'PHW', 'NBA', 'CHS', 'NBA', 4, 1)
(2, 1946, 'QF', 'M', 'NYK', 'NBA', 'CLR', 'NBA', 2, 1)
(3, 1946, 'QF', 'M', 'PHW', 'NBA', 'STB', 'NBA', 2, 1)
(4, 1946, 'SF', 'N', 'PHW', 'NBA', 'NYK', 'NBA', 2, 0)
(5, 1946, 'SF', 'N', 'CHS', 'NBA', 'WSC', 'NBA', 4, 2);
## The user has asked:
What is the difference in the average age of players when they are drafted in the ABA vs when they are drafted in the NBA between the years 1970 and 1970?
NOTE: ABA refers to lgID = 'ABA'; NBA refers to lgID = 'NBA'; between the years 1970 and 1970 refers to draftYear between 1970 and 1970; difference = subtract(avg(subtract(1970, year(birthDate)) where lgID = 'ABA'), avg(subtract(1970, year(birthDate)) where lgID = 'NBA'))
\end{lstlisting}

\begin{lstlisting}[label={lst:correctionprompt}, caption={Error correction prompt for Dubo-SQL}, breaklines=true]

That SQL produced this error message: \"{error_message}\". Write a new query. If you received a 'no such column' error, consider whether you pulled the column from the correct table. Don't apologize. Respond only with SQL.
\end{lstlisting}

\begin{lstlisting}[label={lst:systemprompt}, caption={System prompt for Dubo-SQL v2}, breaklines=true]
You are a principal data engineer. Help users write SQL queries for a SQLite database to answer their questions.

For every user question, you'll be provided with context about their database in the following format:
<table> (
    <column> <data_type> -- <column_description> -- <example_value>, <example_value>, ...
    ...
    <column> <data_type> -- <column_description> -- <example_value>, <example_value>, ...
    FOREIGN KEY (<column>) REFERENCES <table> (<column>)
)

Not all columns have descriptions or example values. Not all tables have foreign keys.

You should only respond in this JSON format:
{
    "sql": "the sql that answers the user's question"
}
\end{lstlisting}

\end{document}